\documentclass[journal]{IEEEtran}
\ifCLASSINFOpdf
\usepackage[pdftex]{graphicx}
\graphicspath{{/figs/}}
\DeclareGraphicsExtensions{.pdf,.eps}
\else
\usepackage[dvips]{graphicx}
\graphicspath{{/figs/}}
\DeclareGraphicsExtensions{.eps}
\fi

\usepackage{xspace}
\usepackage{caption}
\usepackage{subcaption}
\usepackage[cmex10]{amsmath}
\usepackage{amssymb} 
\usepackage{cite}
\usepackage{bm}
\usepackage{algorithm,tabularx} 
\usepackage[noend]{algpseudocode}
\makeatletter
\newcommand{\multilines}[1]{%
	\begin{tabularx}{\dimexpr\linewidth-\ALG@thistlm}[t]{@{}X@{}}
		#1
	\end{tabularx}
}
\makeatother

\usepackage{algpseudocode}
\usepackage{array}

\usepackage{amsthm}
\usepackage{multirow}
\usepackage{color}
\usepackage{epstopdf}
\usepackage{endnotes}
\usepackage{framed}
\usepackage{cool} 
\usepackage{enumerate} 
\usepackage{url}
\usepackage[subnum]{cases}
\usepackage{etoolbox}

\usepackage{lipsum}

\newcommand\blfootnote[1]{%
	\begingroup
	\renewcommand\thefootnote{}\footnote{#1}%
	\addtocounter{footnote}{-1}%
	\endgroup
}

\newcommand{\eqdef}{\mathrel{\mathop=}:}

\begin{document}
	%
	\title{Distributed and Democratized Learning: Philosophy and Research Challenges}
	%
	%
	%
	
	\author{~Minh~N.~H.~Nguyen,~Shashi~Raj~Pandey,~Kyi~Thar,\\~Nguyen~H.~Tran,\IEEEmembership{~Senior~Member,~IEEE},~Mingzhe~Chen,\IEEEmembership{~Member,~IEEE},\\
		~Walid~Saad,\IEEEmembership{~Fellow,~IEEE},~and~Choong~Seon~Hong,\IEEEmembership{~Senior~Member,~IEEE}.
		
		\IEEEcompsocitemizethanks{	
			
			\IEEEcompsocthanksitem M.~N.~H.~Nguyen~,~S.~R.~Pandey,~K.~Thar,~and~C.~S.~Hong are with the Department of Computer Science and Engineering, Kyung Hee University, Yongin-si 17104, South Korea. Email: \{minhnhn, shashiraj, kyithar, cshong\}@khu.ac.kr.
			\IEEEcompsocthanksitem N.~H.~Tran is with School of Computer Science, The University of Sydney, Sydney, NSW 2006, Australia. Email: nguyen.tran@sydney.edu.au.
			\IEEEcompsocthanksitem M.~Chen is with Department of Electrical	Engineering, Princeton University, Princeton, NJ, 08544, USA. Email: mingzhec@princeton.edu.
			\IEEEcompsocthanksitem W.~Saad is with the Wireless@VT, Bradley Department of Electrical and Computer Engineering, Virginia Tech, Blacksburg, VA, 24060, USA. Email: walids@vt.edu.
			
		}

		\thanks{
	}}
	
	\markboth{Preprint}{}%
	
	\maketitle
	\blfootnote{Corresponding Author: Choong Seon Hong (cshong@khu.ac.kr)} 
	\begin{abstract}
		Due to the availability of huge amounts of data and processing abilities, current artificial intelligence (AI) systems are effective in solving complex tasks. However, despite the success of AI in different areas, the problem of designing AI systems that can truly mimic human cognitive capabilities such as artificial general intelligence, remains largely open. Consequently, many emerging cross-device AI applications will require a transition from traditional centralized learning systems towards large-scale distributed AI systems that can collaboratively perform multiple complex learning tasks. In this paper, we propose a novel design philosophy called \emph{democratized learning} (Dem-AI) whose goal is to build large-scale distributed learning systems that rely on the self-organization of distributed learning agents that are well-connected, but limited in learning capabilities. Correspondingly, inspired by the societal groups of humans, the specialized groups of learning agents in the proposed Dem-AI system are self-organized in a hierarchical structure to collectively perform learning tasks more efficiently. As such, the Dem-AI learning system can evolve and regulate itself based on the underlying duality of two processes which we call \emph{specialized} and \emph{generalized processes}. In this regard, we present a reference design as a guideline to realize future Dem-AI systems, inspired by various interdisciplinary fields. Accordingly, we introduce four underlying mechanisms in the design such as \textit{plasticity-stability transition mechanism, self-organizing hierarchical structuring, specialized learning,} and \textit{generalization}.
		Finally, we establish possible extensions and new challenges for the existing learning approaches to provide  better scalable, flexible, and more powerful learning systems with the new setting of Dem-AI.
	\end{abstract}
	

	%
	\IEEEpeerreviewmaketitle
	
	\section{Introduction}	
	The growing success of AI in real-life applications has proliferated its usage. AI has provided a plethora of solutions for complex problems across multiple fields such as decision support systems in healthcare, automation in retail and industries, advanced control and operations, and telecommunications, among others. Correspondingly, numerous research activities in machine learning technologies \cite{Chen_tutorial, FL_advances, tran2019federated, pandey2020crowdsourcing, FL_TON2019, khan2019federated, chen2019joint,chen2020convergence} focused on architectures and algorithm designs that empowered the emergence of cross-device AI applications in our daily lives. However, in practice, the performance efficiency and re-usability of trained AI systems are quite limited, particularly when seeking to solve multiple complex learning tasks and when dealing with unseen data, due to their rigid design and learning settings. To address these issues, the recently proposed meta-learning framework (MLF) \cite{maml2017} provides capabilities that allow generalization from large training of similar tasks. Hence, MLF is able to quickly adapt to similar new tasks using only a small number of training samples. Meanwhile, the so-called multi-task learning (MTL) frameworks introduced in \cite{mocha2017} and \cite{corinzia2019variational} allow training a general model for multiple small number of tasks; however, it requires significant similarity among those tasks. Therefore, there is an imminent need for rethinking existing machine learning systems and transforming them into systems that can control the generalization ability (i.e., good performance on unseen data of a single/multiple tasks) together with specialization ability (i.e., good performance on the learning tasks). 
	
	\begin{figure*}[t]
		\centering
		\includegraphics[width=\linewidth]{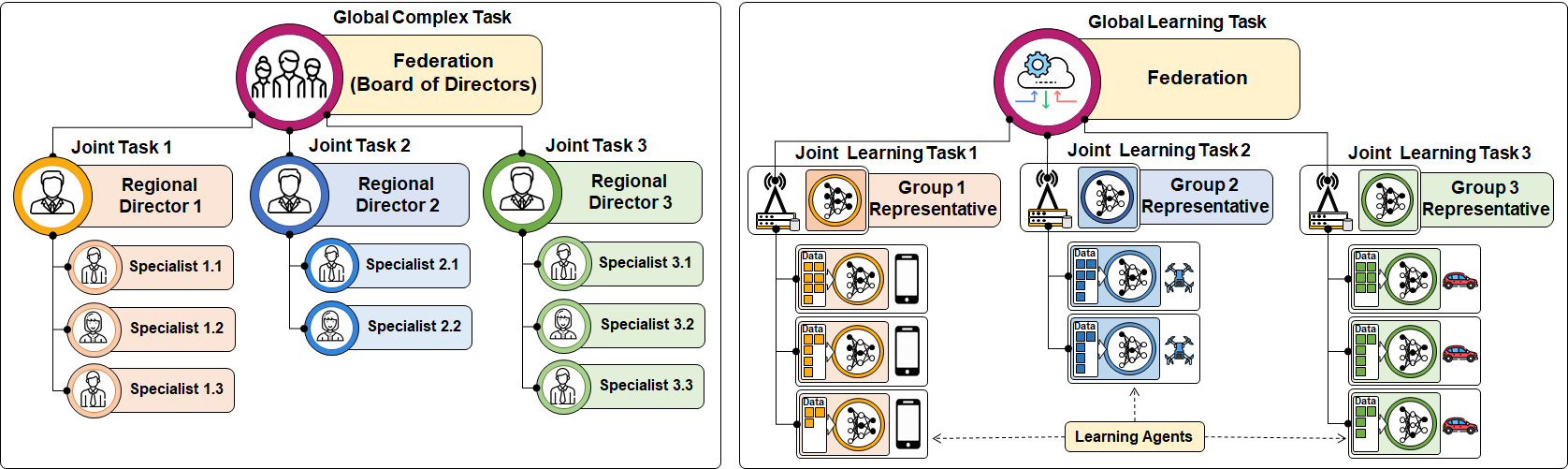}
		\caption{Analogy of a hierarchical distributed learning system.}
		\label{F:Motivation}
	\end{figure*}

	\subsection{Towards a Large-scale Distributed Learning System} 
	
	AI is moving towards edge devices with the availability of massively distributed data sources and the increase in computing power for handheld and wireless devices such as smartphones or self-driving cars. This has generated a growing interest to develop large-scale distributed machine learning paradigms \cite{corinzia2019variational}. In this regard, the edge computing paradigm provides the underlying infrastructure that empowers regional learning or device learning at the network's edge. However, traditional learning approaches cannot be readily applied to a large-scale distributed learning system. One promising approach to build a large-scale distributed learning system is through the use of the emerging federated learning (FL) framework \cite{FL_advances}. In FL, on-device learning agents collaboratively train a global learning model without sharing their local datasets. The global model at the central server allows the local model to improve the learning performance of each agent; however, iteratively updating the global model based on the aggregation of local models can also have a negative impact on the personalized performance \cite{FL_advances}. For example, in a supervised FL setting, the local model is optimized to fit the local dataset, whereas the global model is built on the simple aggregation of local learning parameters (e.g., FedAvg \cite{mcmahanCommunicationEfficientLearningDeep2017}, FedProx \cite{FedProx2020}) so as to perform well on the distributed dataset. In practice, local datasets collected by each agent are unbalanced, statistically heterogeneous, and exhibit non-i.i.d (non-independent and non-identically distributed) characteristics. Thus, the global model of FL can become biased and strongly affected by the agents who have more data samples or by those who perform larger update steps during the aggregation of local model parameters. Consequently, beyond a certain threshold value of training rounds, the generalized global model can negatively affect the personalized performance of several learning agents \cite{FL_advances}.
	Hence, the conventional FL cannot efficiently handle the underlying cohesive relation between the generalization and personalization (or specialization) abilities of the learning model in the testing and validation phase \cite{FL_advances}. This raises an important, fundamental research question: \textit{How can one resolve the discrepancies between global and personalized accuracy?} 
	To answer this question and overcome the aforementioned limitations of existing FL frameworks, we seek to develop a novel design philosophy which can be widely used for future large-scale distributed learning systems. 
	To the best of our knowledge, the work in \cite{fallah2020personalized} was the first attempt to study and improve the personalized performance of FL using a so-called personalized federated averaging (Per-FedAvg) algorithm based on MLF. However, in \cite{fallah2020personalized} the cohesive relation between the generalization and personalization were not adequately analyzed. Recent work in \cite{mansour2020three} developed an analysis for the personalization in FL applications using three approaches, such as hypothesis-based clustering, data interpolation, and model interpolation.

	%
	
	
	\subsection{Lifelong Learning and Formation of a Hierarchical Structure}\label{SS:Lifelong learning and formation fo a social structure}
	Inspired from the lifelong learning capability of biological intelligence and systems \cite{LifeLong2019}, we observe that both generalization and specialization capabilities are involved in building a large-scale distributed learning system. As is the case in any form of biological intelligence, we observe a continual developmental process: from stem cells to complex structures with multiple functionalities, such as the human brain. For example, in humans, the learning process consists of many stages such as \textit{newly born}, \textit{childhood}, and \textit{grown-up}. A \textit{newly born} stage characterizes the \textit{generalization capability} with a high level of neurosynaptic \emph{plasticity} \cite{LifeLong2019} in a human brain. The synaptic plasticity level is intrinsically involved to consolidate knowledge for learning and adapting to the dynamic environment. In this learning stage, an individual can vastly learn basic functions/skills and abilities under the influence of social adaptation and education, or by leveraging curiosity, implicit or explicit rewards \cite{oudeyer2018computational}. The transition from the \textit{newly born} stage to the \textit{childhood} stage is characterized by the individual's pursuit to have a \textit{specialization capability} over a set of already known basic skills and to further explore the world with a hierarchical structure of generalized knowledge which can help to perform the complex tasks. At the \textit{grown-up} stage, individuals are more able to efficiently deal with highly complex tasks, i.e., better adaptation ability in the known environment for solving complex learning tasks. However, they also lose the power of generalization capabilities, i.e., it is harder for those individuals to learn new things due to the improvement of knowledge consistency following various developmental stages \cite{LifeLong2019}. 
	
	\begin{figure*}[t]
		\centering
		\includegraphics[width=.85\linewidth]{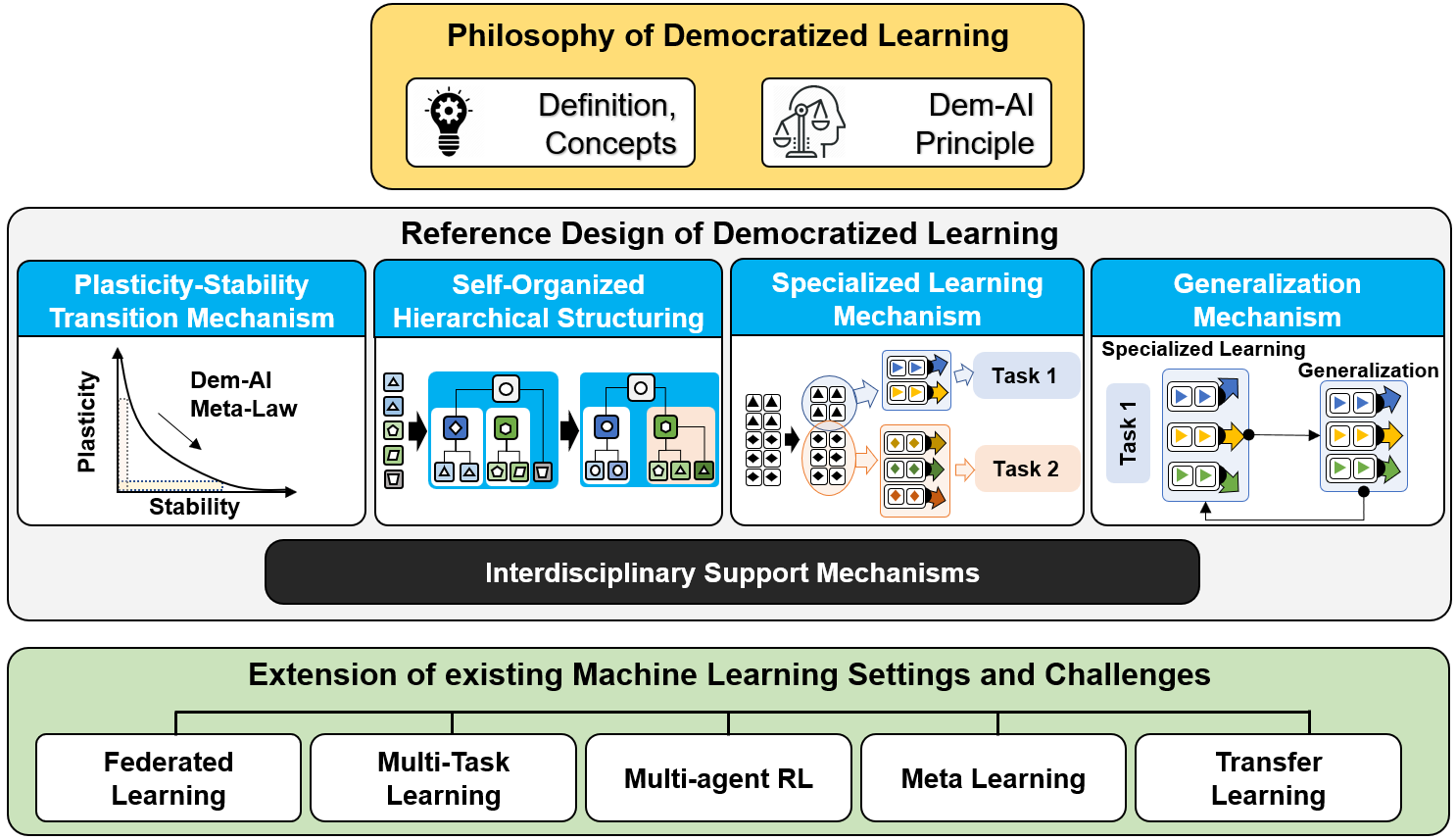}
		\caption{Anatomy of Democratized Learning.}
		\label{F:Building_Blocks}
	\end{figure*}
	
	In another observation regarding the role of individual capabilities in the social structure development process, individuals contribute to society by resolving multiple complex tasks in a way similar to the learning agents in large-scale distributed learning systems. An individual exists as a unit entity with some basic survival objectives and functions/skills in the social hierarchy while \textit{interacting, contributing,} and \textit{forming smaller groups} such as a family. The conglomeration of families and relatives characterizes a society that behaves as a bigger group to resolve complex life issues and resolve/create conflicts. Subsequently, a union of such groups within a border demarcation represents a state which has their own distinct legal regulations and social structures to solve complex social issues. The states form a global world organization, such as the \textit{United Nations}, to maintain global harmony and solve overly complex global issues. Thus, many small groups unite to form a hierarchical structure for knowledge sharing and solving complex tasks. Here, the group formation process is a common purpose of shared benefits among the members or the smaller social groups. Moreover, the structure analogous to human society results in the collective behavior of an interactive crowd, often characterized by \textit{swarm intelligence}, which is well-observed in numerous biological systems \cite{brambilla2013swarm}. Over time, such collaboratively-built social structures become stable and more consistent. Fig.~\ref{F:Motivation} illustrates an analogy between the hierarchical structure in organizations (companies) and a hierarchical distributed learning system. From this figure, we can see that a global complex task can be accomplished through the cooperation of outcomes at each division. Following a similar analogy, a global learning task can be solved through the collaboration of each individual group's learning outcomes.

	These observations provide sufficient hints about the underlying duality of the generalized and specialized processes in the entire development process of biological intelligence or systems. These processes eventually integrate many basic skills to behave/become complex skills; they start from generalized knowledge at a high level of plasticity and head towards more specialized ones at a high level of stability. This also raises an important question: \textit{How can one understand and formalize the duality of the generalized and specialized processes regarding the \textit{plasticity} and \textit{stability} of a distributed learning system}?
	
	\subsection{Our Contributions}
	
	Existing distributed AI systems such as FL focus on building the federation of a central server and clients to construct a global model from the aggregation of personalized models, irrespective of the differences in agents' capabilities and the characteristics of their local dataset and learning task. Different from FL, we consider a self-organizing learning framework with a hierarchical structure for solving multiple complex learning tasks. Furthermore, considering the observations in Section \ref{SS:Lifelong learning and formation fo a social structure}, we develop a novel philosophy to analyze the following research questions for collective learning using a large number of biased learning agents, who are committed to learn from limited datasets and learning capabilities:
	\begin{itemize}
		\item How can large-scale distributed AI systems be self-organized in a suitable hierarchical structure to perform knowledge sharing?
		\item How can learning knowledge be shared among learning agents and tasks?
		\item How can learning agents integrate the generalized knowledge to enhance their learning performance?
	\end{itemize}
	Learning in each agent can introduce bias due to the following scenarios: a) limited number of samples in the local data, b) inadequate availability of features in the local data, c) unbalanced data (e.g., heterogeneity of labeled data in classification problem \cite{FL_advances}), or d) limited information (e.g., partial observation) regarding the environments \cite{NIPS2019_multiagent}. In general, the direct consequence of all of these scenarios is that they produce biased personalized knowledge for each learning agent, which we refer to as the \emph{``biased agent"}. Thus, a collaboration between agents is required to improve the generalized learning performance of all agents.
	
	\begin{figure*}[t]
		\centering
		\includegraphics[width=.8\linewidth]{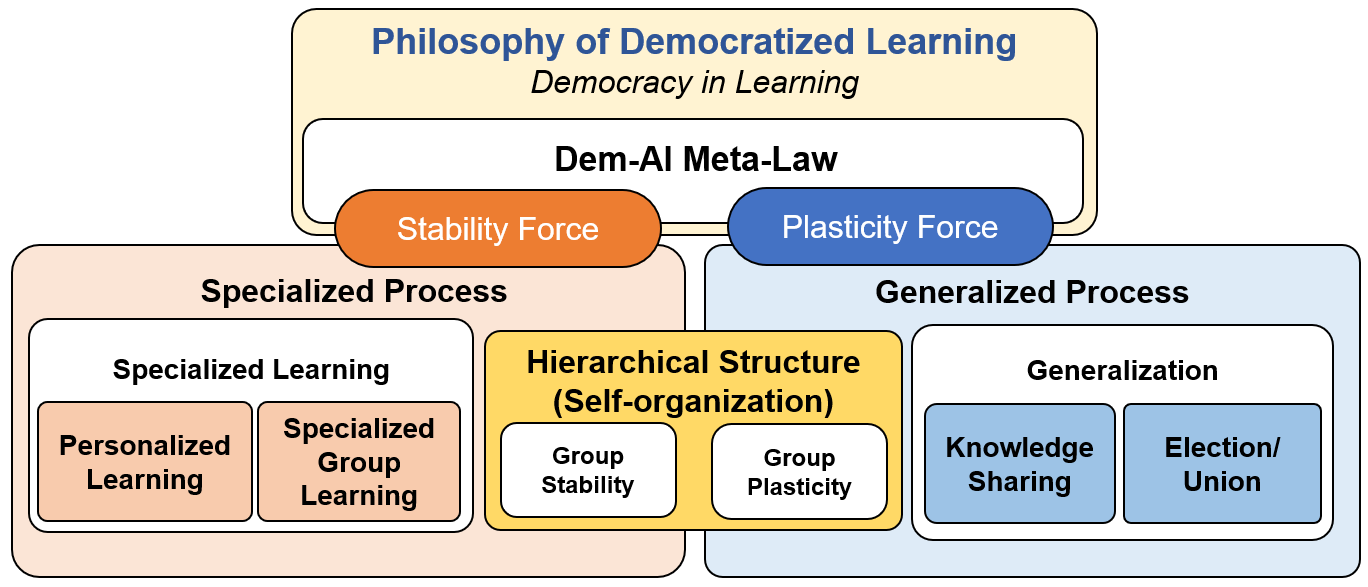}
		\caption{Conceptual architecture of the proposed democratized learning philosophy.}
		\label{F:DemAI}
	\end{figure*}
	
	In a nutshell, our key contribution is to develop fundamental principles for building large-scale distributed AI systems from the self-organizing hierarchical structures that consist of a very large number of biased individuals (learning agents). In our envisioned system, voluntary contributions from the learning agents enable collaborative learning that empowers the hierarchical structure to solve multiple complex tasks while supporting each agent to improve its learning performance. Furthermore, this collaboration will speed up the learning process for the new members without prior learning knowledge and add benefits in expanding the generalized knowledge of the groups. Accordingly, we study the underlying dual specialized-generalized processes to develop a ``philosophy and design" of a new distributed learning system, namely \textit{democratized learning} or \textit{Dem-AI}, in short. Furthermore, the design philosophy of Dem-AI opens up many new research challenges and extensions of the existing learning settings for FL, MTL, MLF, multi-agent reinforcement learning (MARL), and transfer learning. To provide a holistic view, we present the anatomy of Dem-AI as shown in Fig. \ref{F:Building_Blocks}. The rest of the paper is organized as follows. 
	
	We introduce the fundamental concepts and principles of the proposed democratized learning philosophy in Section \ref{S: DemAI Philosophy}. A reference design of democratized learning is presented in Section \ref{S: Reference Design of Democratized Learning}, followed by the possible extension of existing machine learning settings in Section \ref{S: Extension of Existing ML Settings and Challenges}. Section \ref{S:Conclusion} concludes the paper.

	\section{Democratized Learning Philosophy}\label{S: DemAI Philosophy}
	In this section, we introduce our Dem-AI philosophy including the definition, the concepts as shown in Fig. \ref{F:DemAI}, and the principle related to the evolution of the self-organizing hierarchical structure, specialized learning, and generalization in the new democratized learning system.

	\subsection{Definitions, Goal, and Concepts}
	\textbf{Definition and goal:} Democratized Learning (\textit{Dem-AI} in short) focuses on the study of dual (coupled and working together) specialized-generalized processes in a self-organizing hierarchical structure of large-scale distributed learning systems. The specialized and generalized processes must operate jointly towards an ultimate learning goal identified as performing collective learning from biased learning agents, who are committed to learn from their own data using their limited learning capabilities.
	
	As such, the \emph{ultimate learning goal} of the Dem-AI system is to establish a mechanism for collectively solving common (single or multiple) complex learning tasks from a large number of learning agents. In case of a common single learning task setting, the Dem-AI system aims to improve the aggregation accuracy of all agents, as done in federated learning. The ultimate goal in a conventional federated learning system is to minimize the average model loss of all agents for a single learning task. Moreover, for different learning settings and applications, the goal of Dem-AI systems can be derived from following specific designs of learning objectives. The learning agents can also collectively contribute to solving common multiple tasks as an ultimate goal in Dem-AI system. For example, with multi-tasks learning setting \cite{ruder2017overview}, the goal is to attain the overall learning performance for multiple tasks, simultaneously. In meta-learning \cite{maml2017}, the learning goal is to construct a generalized knowledge that can efficiently deal with similar new learning tasks. Similarly, in reinforcement learning, the coordination task is defined to maximize the cumulative rewards for the joint actions taken by individual agents according to their partial observation \cite{NIPS2019_multiagent}.

	\textbf{Democracy in learning:} 
	The Dem-AI system features a unique characterization of participation in the learning process, and consequently develops the notion of \emph{democracy in learning} whose principles include the following:
	\begin{itemize}
		\item According to the differences in their characteristics, learning agents are divided into suitable groups that can be specialized for the learning tasks. These specialized groups are self-organized in a hierarchical structure to mediate \emph{voluntary contributions} from all members in the collaborative learning for solving multiple complex tasks. 
		\item The shared generalized learning knowledge supports specialized groups and learning agents to improve their learning performance by reducing individual biases during participation. In particular, the learning system allows new group members to: a) speed up their learning process with the existing group knowledge and b) incorporate their new learning knowledge in expanding the generalization capability of the whole group.
		\item Learning agents are free to join suitable learning groups and exhibit equal power in constructing their groups’ generalized learning model. The group power can be represented by the number of its members, which can vary over the training time.
	\end{itemize}
	
	\textbf{Dem-AI Meta-Law:} We define a \emph{meta-law} as a mechanism that can be used to manipulate the transition between the dual specialized-generalized processes of our Dem-AI system. This meta-law is driven by two coincident primary forces: 1) a \emph{stability force}, and 2) a \textit{plasticity force}. Throughout the learning time, the transition mechanism adjusts the importance weight between these forces to empower the plasticity or the stability in the specialized learning and generalization as well as the hierarchical structure of the Dem-AI system. The Dem-AI meta-law also provides the necessary information to regulate the self-organizing hierarchical structuring mechanism.   
	
	\textbf{Specialized Process:} This process is used to leverage the \textit{specialized learning} capabilities in the learning agents and specialized groups by exploiting their collected data. This process also drives the hierarchical structure of specialized groups with many levels of relevant generalized knowledge to become stable and well-separated. Thus, with the addition of higher levels of generalized knowledge created by the generalization of all specialized group members, the learning agents can exploit their local datasets so as to reduce biases during \emph{personalized learning} for a single learning task. Thus, the personalized learning objective has two goals in its learning problem: 1) to perform \emph{specialized learning}, and 2) \emph{to reuse the available hierarchical generalized knowledge}. 
	Besides, generalized knowledge can be incorporated by the regularizers for the personalized learning objectives.
	Moreover, the specialized learning can be performed as group learning formation when the members do not have learning capabilities on their own or when they are required to solve a coordination task. 
	Considerably, over time, the generalized knowledge becomes less important compared to the specialized learning goal and a more stable hierarchical structure of specialized groups will form. These transitions are the direct consequence of the \textit{stability force} characterized by the specialized knowledge exploitation and knowledge consistency. However, it becomes stronger over time in the meta-law design of our Dem-AI principle. 
	
	\begin{figure*}[t]
		\centering
		\includegraphics[width=\linewidth]{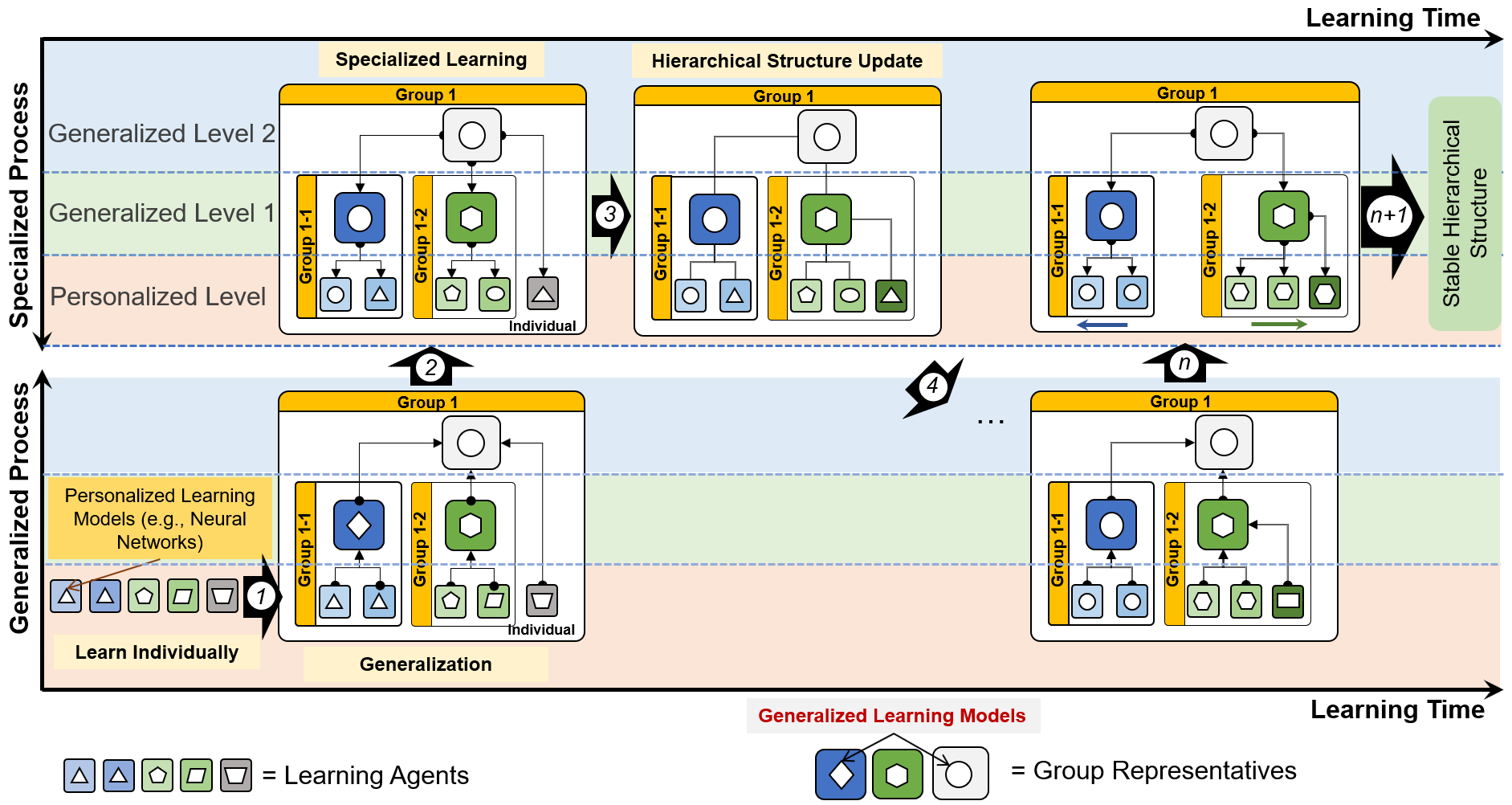}
		\caption{The illustration of the transition in Dem-AI principle.}
		\label{F:Dem_AI_Principle}
	\end{figure*}
	
	
	\textbf{Generalized Process:} This process is used to regulate the \emph{generalization} mechanism for all existing specialized groups as well as the plasticity level of all groups. Here, \emph{group plasticity} pertains to the ease with which learning agents can change their groups. The generalization mechanism encourages group members to become close together when doing a similar learning task and sharing knowledge. 
	This process enables the sharing of knowledge among group members and the construction of a hierarchical level of the generalized knowledge from all of the specialized groups. Thereafter, the generalized knowledge helps the Dem-AI system to maintain the generalization ability for efficiently dealing with environment changes or new learning tasks. Hence, \emph{knowledge sharing} is the mechanism to construct the generalized knowledge from similar and correlated learning tasks such as model averaging in FL \cite{mcmahanCommunicationEfficientLearningDeep2017}, sharing knowledge mechanisms in multi-tasks learning \cite{zhang2017survey}, and knowledge distillation \cite{park2019distilling}. Moreover, to resolve the conflict among excessively different specialized groups, an \textit{election} mechanism can be adopted to reach consensus or a \textit{union} mechanism can be applied to maintain the diversity of the potential groups. Consequently, the hierarchical generalized knowledge can be constructed based on the contribution of the group members, which is driven by the \textit{plasticity force}. This force is characterized by creative attributes, knowledge exploration, multi-task capability, and survival in uncertainty, and it becomes weaker over time in the meta-law design of the Dem-AI principle.	
	
	\textbf{Self-organizing Hierarchical Structure:}
	According to the transition between the two basic forces as well as the necessary information in the meta-law, the hierarchical structure of specialized groups and the relevant generalized knowledge are constructed and regulated following a self-organization principle (e.g., hierarchical clustering \cite{H_Clustering}). This structure then evolves to become more stable: temporary small groups with high-level group plasticity will later unite to form a bigger group that enhances the generalized capability for all members.
	Thus, the specialized groups at higher levels in the hierarchical structure have more members and can construct more generalized (less biased) knowledge. Hierarchical modular networks can be found in the human brain as well as in the structures of human knowledge \cite{zurn2020networkhierarchy}. These hierarchical structures exhibit higher overall performance and evolvability (i.e., faster adaptation to new environments), as explained in \cite{mengistu2016evolutionaryhierarchy}.
	
	Next, we establish the general principles that characterize the evolution of underlying processes in Dem-AI.
	
	\subsection{Dem-AI Principle}
	\textbf{Transition in the dual specialized-generalized process during the training:}
	Throughout the learning time, the specialized process becomes dominant over the generalized process to perform better in the training environment following the Dem-AI meta-law design. This transition induces the following evolution principles of Dem-AI
	\begin{itemize}
		\item \textbf{P1: Evolution of specialized learning and generalization:}
		The transition due to the duality of the two processes keeps the Dem-AI system evolving in order to provide a better adaptation ability for solving complex learning tasks during training. The Dem-AI system observes an incremental impact of the specialized learning over the learning time and also loses the power of generalization, i.e., a decremental opportunity to deal with environment changes, such as unseen data, new learning agents, and new learning tasks.
		
		\item \textbf{P2: Evolution of the self-organizing hierarchical structure:}
		The transition due to the duality of the two processes keeps the self-organizing hierarchical structure of the Dem-AI system evolving from a high level of plasticity to a high level of stability, i.e., from unstable specialized groups to well-organized and well-separated specialized groups. 
	\end{itemize}
	In this transition, the separation of the specialized groups at each level is accelerated as a consequence of (\textbf{P1}), thereby increasing the resistance of learning agents to change their groups. Meanwhile, the evolution of the self-organizing hierarchical structure (\textbf{P2}) accelerates the evolution of specialized learning and generalization (\textbf{P1}). 
	
	\begin{figure*}[t]
		\centering
		\begin{subfigure}{0.36\linewidth}
			\includegraphics[width=0.9\linewidth]{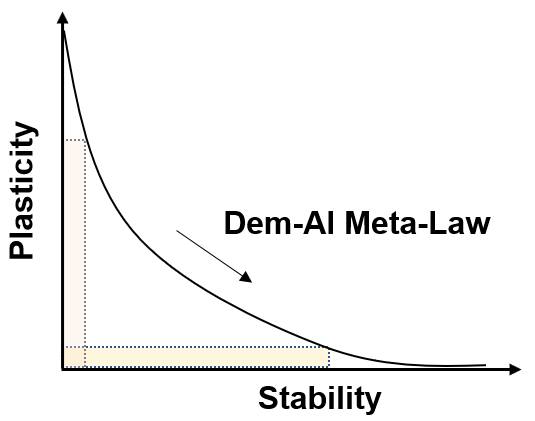}
			\caption{The transition between plasticity and stability.}
			\label{F:Transistion}
		\end{subfigure}
		\begin{subfigure}{0.62\linewidth}
			\includegraphics[width=\linewidth]{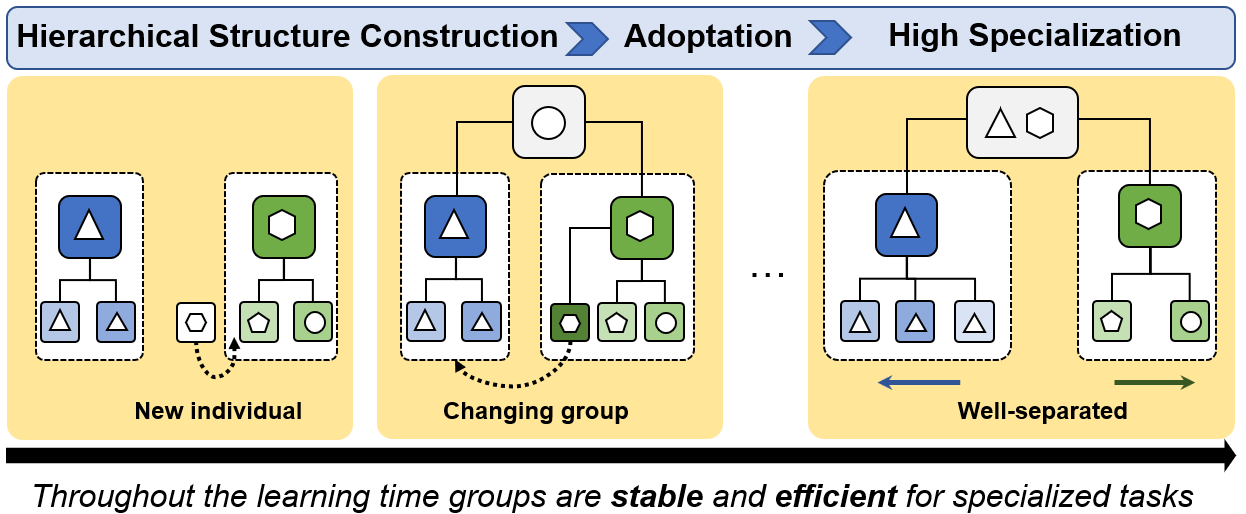}
			\caption{Self-organizing hierarchical structuring.}
			\label{F:Hierarchical_Structuring}
		\end{subfigure}
		\caption{The transition between plasticity and stability and the self-organizing hierarchical structuring in the Dem-AI system.}
	\end{figure*}
	\setlength{\textfloatsep}{7pt}

	Throughout the training process, predefined goals such as maximizing the rewards or minimizing learning loss enables the learning agents to attain higher performance in the fixed training environment. Learning agents therefore gain specialized capabilities for the fixed training environment and eventually reduce their generalized capability to adapt to changes in the applied environments. In this regard, the Dem-AI principle hypothesizes the transitions between (\textbf{P1}) and (\textbf{P2}), which helps the learning agents to obtain better learning goals in the training process of Dem-AI systems. We then realize the mechanism defined as (\textbf{P1}) by controlling the parameters in the specialized learning objective of learning agents and generalization in the next section. The coherence between \emph{group plasticity} is based on the stability of specialized groups and the learning efficiency of specialized learning. In (\textbf{P1}), the group plasticity accelerates the separation of the specialized groups at each level due to the difference in their learning characteristic (e.g., by updating models towards different learning directions). Thus, this process increases the resistance of learning agents to change their groups. To speed up (\textbf{P2}), the Dem-AI system can also directly control the resistance of changes in the self-organizing hierarchical structuring mechanism.
	
	The transition following the Dem-AI principle in the dual specialized-generalized process is illustrated in Fig. \ref{F:Dem_AI_Principle}. In this transition, the learning agents are grouped according to the similarities of their learning tasks at the early stage. Then, the generalized process helps in the construction of the hierarchical generalized knowledge for the specialized groups and encourages the group members to be close together. In the meantime, the specialized learning processes leverage the personalized and specialized group learning to exploit their biased dataset and deviate from the common generalized knowledge. 
	The hierarchical structure becomes stable with the coexistence of well-separated highly complex specialized groups to provide different highly efficient specialized models for solving the complex learning tasks. This may, however, lead to ``overfitting" in the training environment. Therefore, to deal with the environment changes we should properly control this dual process to achieve a high specialized performance while preserving the generalized capabilities of the Dem-AI learning system.
	
	\section{Reference Design of Democratized Learning}\label{S: Reference Design of Democratized Learning}
	In the previous section, we introduced the fundamental concepts and general principles in the Dem-AI philosophy for democratized learning. In this section, we initiate a reference design with guidelines for the Dem-AI philosophy that is inspired by observations of various interdisciplinary mechanisms in nature. Specifically, a Dem-AI system requires four essential mechanisms: \textit{transition mechanism of plasticity-stability, self-organizing hierarchical structuring, specialized learning,} and \textit{generalization}, which will be presented in the following subsections.

	
	
	
	\subsection{Plasticity-Stability Transition Mechanism in Dem-AI Meta-Law}
	The transition of the plasticity and stability of Dem-AI systems in the meta-law design can drive the evolution of the specialized and generalized processes, following suitable mechanisms based on the characteristic of the learning systems. As shown in Fig. \ref{F:Transistion}, according to the Dem-AI principle, the specialized process with the incremental stability force becomes dominant in the generalized process with the decremental plasticity force. To implement this transition, we can approximate the whole learning process by different stages that change from a high level of plasticity to a high level of stability in specialized learning, generalization, and the self-organizing hierarchical structuring mechanism.
	Specifically, the meta-law can be designed and operated as a global rule at the global controller for the whole system. However, decentralization of the learning process requires a design that adds flexibility in controlling the parameters for the generalization (e.g., $\gamma^t$) and specialization learning mechanisms (e.g., $\alpha^t,\beta^t $) at the group level or device level, which are introduced in the next subsections. 
	This way, we can avoid fixed global parameters in the meta-law which is applied to all of the learning agents and groups. Furthermore, these controllable parameters can depend on how long the groups are created and how long the agents participate in the system.
	
	Analogously, in physics, we associate the transition in our meta-law with the pendulum principle \cite{pendulum} that shows a transition from the potential to the kinetic energy by the energy conversation law and exhibits a cyclic increment or decrement in sine forms. This analogy additionally reveals a hidden relationship between stability and plasticity that can inspire other suitable engineering mechanisms.
	Thereafter, we can incorporate Hebbian and homeostatic plasticity mechanisms, studied extensively in neuroscience \cite{LifeLong2019}, to regulate the Dem-AI systems.
	
	\begin{figure*}[t]
		\centering
		\begin{subfigure}{0.61\linewidth}
			\includegraphics[width=0.9\linewidth]{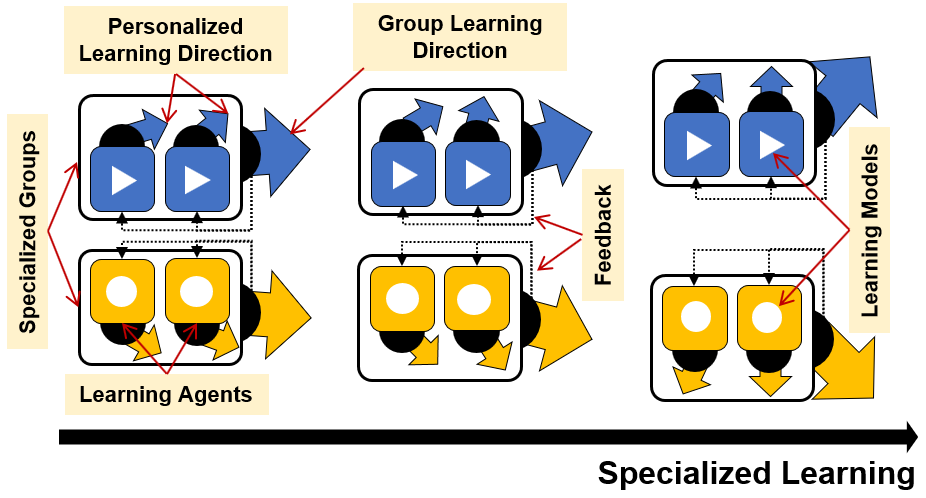}
			\caption{Specialized learning mechanism.}
			\label{F:Specialized_Learning}
		\end{subfigure}
		\begin{subfigure}{0.37\linewidth}
			\includegraphics[width=\linewidth]{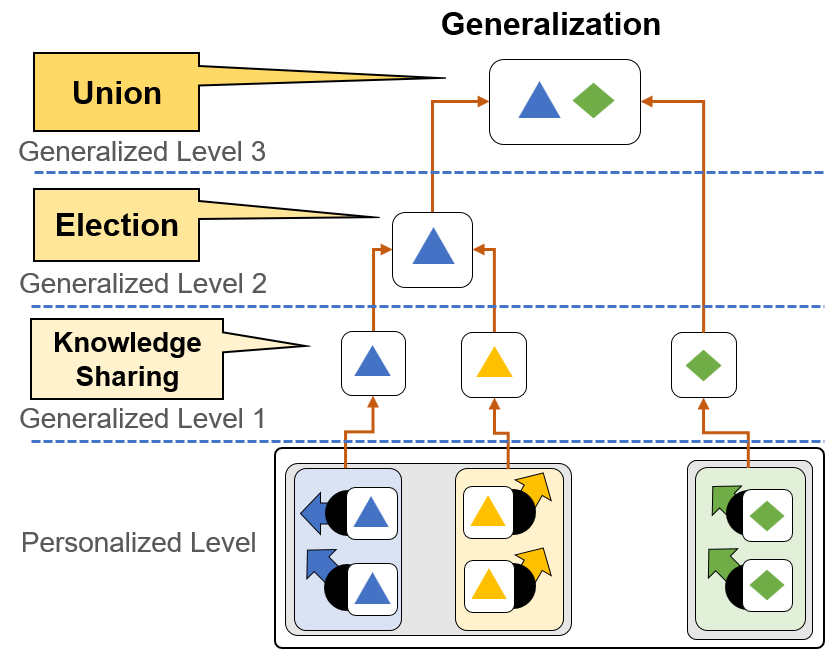}
			\caption{Hierarchical generalization mechanism.}
			\label{F:Generalization}
		\end{subfigure}
		\caption{Specialized learning and generalization mechanism in the Dem-AI system.}

	\end{figure*}

	\subsection{Self-Organizing Hierarchical Structuring Mechanism}
	Fig. \ref{F:Hierarchical_Structuring} shows the self-organizing hierarchical structuring mechanism that helps in constructing and maintaining the structure of many levels of specialized groups. 
	This structuring process can be divided into three stages:
	\begin{itemize}
		\item \textbf{Early-Stage (Hierarchical structure construction):} The lowest-level groups are created by grouping the agents who perform a common learning task and have similar characteristics in their learning models. A new level can be created when the measure of distances among current groups is greater than a threshold pre-defined for each level. The structure can reach the maximum number of levels which is defined in the Dem-AI meta-law. Alternatively, we can extend existing hierarchical clustering algorithms \cite{H_Clustering}, clustering mechanisms for FL \cite{sattler2020clustered, mansour2020three} or game-theoretic mechanisms \cite{han_niyato_saad_basar_2019,hierarchicalgame_2014,coalitionalgame}.
		\item \textbf{Adaptation Stage}: The adaptation stage allows the learning agents to change their groups. When the level of group plasticity is high, the measured distances among specialized groups are short, and, as such, agents can move among these groups.
		\item \textbf{High Specialization Stage}: The Dem-AI system allows micro-adjustments in low-level groups due to the well-separated and stable specialized groups that are already formed.
	\end{itemize}
	
	New learning agents first join in a suitable specialized group in the top-level of the learning system. In time, these agents will be admitted to lower-level specialized groups with whom they share similarities in the learning characteristics. 
	Consequently, these agents leverage the existing group knowledge to speed up their learning process. Furthermore, with the availability of new data, the agents can contribute their valuable personalized knowledge to improve the generalization capability of the groups. Note that the metric of measured distance in the structuring mechanism can be derived from the differences in the characteristics of learning agents or specialized groups (e.g., in FL, the metric can be a Euclidean distance of model parameters, gradients, or momentum of learning agents or groups). Therefore, the policy for the system to group different agents and change groups can be defined by the threshold metrics, measured as the differences between the learning agents and between the groups. The recent work in \cite{kornblith2019similarity} provides a promising approach to analyze the similarity of layers in the neural network representation based on centered kernel alignment (CKA).
	
	In swarm intelligence system, e.g., swarm robotics, the self-organized behaviors of a large number of robots can coordinate with each other to design robust, scalable, and flexible collective behaviors \cite{brambilla2013swarm}, which can be instrumental for our mechanism design. 
	Similarly, in addition to the development process of a social structure, the other suitable mechanisms such as the growth process in a biological cell can be incorporated. In biology, the solid complex composited structures such as DNA can be separated after the initial coincidence period in the cell division process. For well-separated groups that are formed, the agents who become excessively different through personalized learning (e.g., different gradient and personalized model parameters) or those who function poorly can be eliminated from their groups. Analogously, such mechanisms can be found in immune responses that destroy unhealthy cells (e.g., cancer or virus-infected cells) \cite{iannello2013immune}. However, we can also consider that these agents behave as new learning agents and move towards other suitable groups.

	\subsection{Specialized Learning Mechanism}
	Specialized learning facilitates the personalized and specialized group learning capability using existing hierarchical generalized knowledge that is represented in Fig. \ref{F:Specialized_Learning}. For this mechanism, we discuss the general design of personalized learning and specialized group learning, as well as related problems.	
	
	\textbf{Personalized learning problem of learning agents:} 
	In Dem-AI, a personalized learning problem can be constructed for each learning agent with a personalized learning objective (\textbf{PLO}) that comprises: 1) a personalized learning goal (\textbf{PLG}), and 2) a reusable generalized knowledge (\textbf{GK}). For example, \textbf{PLG} is the learning loss function, and \textbf{GK} is the regularizer defined as the difference between the new model parameters and the model parameters of the higher-level specialized groups (e.g., FEDL \cite{FL_TON2019}, FedProx \cite{FedProx2020}), i.e.,
	\begin{align}
		\textbf{PLO} 
		&=  \alpha^t\textbf{PLG} + \beta^t \textbf{GK} \nonumber\\
		&=  \alpha^t\textbf{PLG} + \beta^t \sum_{k}\frac{1}{N_g^{(k)}}\textbf{GK}(N_g^{(k)}),
	\end{align}
	where $N_g^{(k)}$ and $\textbf{GK}(N_g^{(k)})$ are the number of agents and the generalized knowledge in the specialized group level $k$, respectively. The higher levels of generalized knowledge are less important when solving any specific learning task of the agents than the lower-level specialized knowledge. Since the specialized process is more important when improving the specialized capability of the personalized learning model, the weight parameter $\beta^t$ must be decreased in order to reduce the plasticity while $\alpha^t$ must be increased according to the meta-law design. If the learning agents cannot directly incorporate the generalized knowledge (e.g., they do not have the same model parameters), a special integration mechanism for the hierarchical structure of knowledge is required. Moreover, computing, communication resource and delay constraints also need to be considered in the learning problem. 
	An example of the specialized learning problem using the proximal term to constrain the local learning model, such that it be closer to the learning model at the higher-level groups, is defined as follows:
	\begin{align}
		\label{EQ:LPL_level_0}
		\underset{\boldsymbol{w}}{\min}~ \alpha^t L^{(0)}_n (\boldsymbol{w}|\mathcal{D}^{(0)}_n) + \beta^t\sum_{k}\frac{1}{N_g^{(k)}}\|\boldsymbol{w}-\boldsymbol{w}_n^{(k)}\|^2;   
	\end{align}
	where $L^{(0)}_k$ is the learning loss function of the learning agent $n$ for a classification task given its personalized dataset $\mathcal{D}^{(0)}_k$ and the learning model ${w}_n^{(k)}$ of the higher-level groups \cite{nguyen2020}. Also, other knowledge transfer techniques such as multi-task regularizer \cite{zhang2017survey} or knowledge distillation regularizer \cite{park2019distilling} can be used to define \textbf{GK} in the \textbf{PLO} of learning agents.
	
	\textbf{Specialized learning problem of specialized groups:}
	In case of group members who do not have adequate learning capabilities (e.g., IoT sensing devices with low computational capabilities), or  learning agents who are required to solve a coordination task. For example, in practical IoT applications, a given learning system may not always be able to guarantee the participation of all clients in every communication round due to intermittent communication, battery drainage, or hardware ailments \cite{sattler2019robust}. Therefore, a specialized group learning problem can be performed at the edge servers, and/or fog nodes at the network's edge as done, for example, in the In-Edge AI framework \cite{wang2019edgeIntelligence}. The goal of this problem is to fit the collective datasets of all group members, where the group behaves as a virtual agent that solves the learning problem. Furthermore, the specialized group learning can also have special decentralized learning structures (e.g., sharing of critic network in multi-agent deep reinforcement learning (DRL) \cite{ma_ddpg2017}, meta-training phase in MLF \cite{maml2017}). Similar to the personalized learning problem, specialized group learning needs to be extended by leveraging the generalized knowledge from the higher-level specialized groups. Next, in order to achieve joint energy-learning efficiency given the limited communication and computation resources, the design of group learning problem needs to consider a synergy of resource allocation, device scheduling, and learning performance. In doing so, the learning system can accommodate large group members and ensure more frequent model updates, thereby improving the group learning performance.
	
	\begin{figure*}[t]
		\centering
		\includegraphics[width=0.81\linewidth]{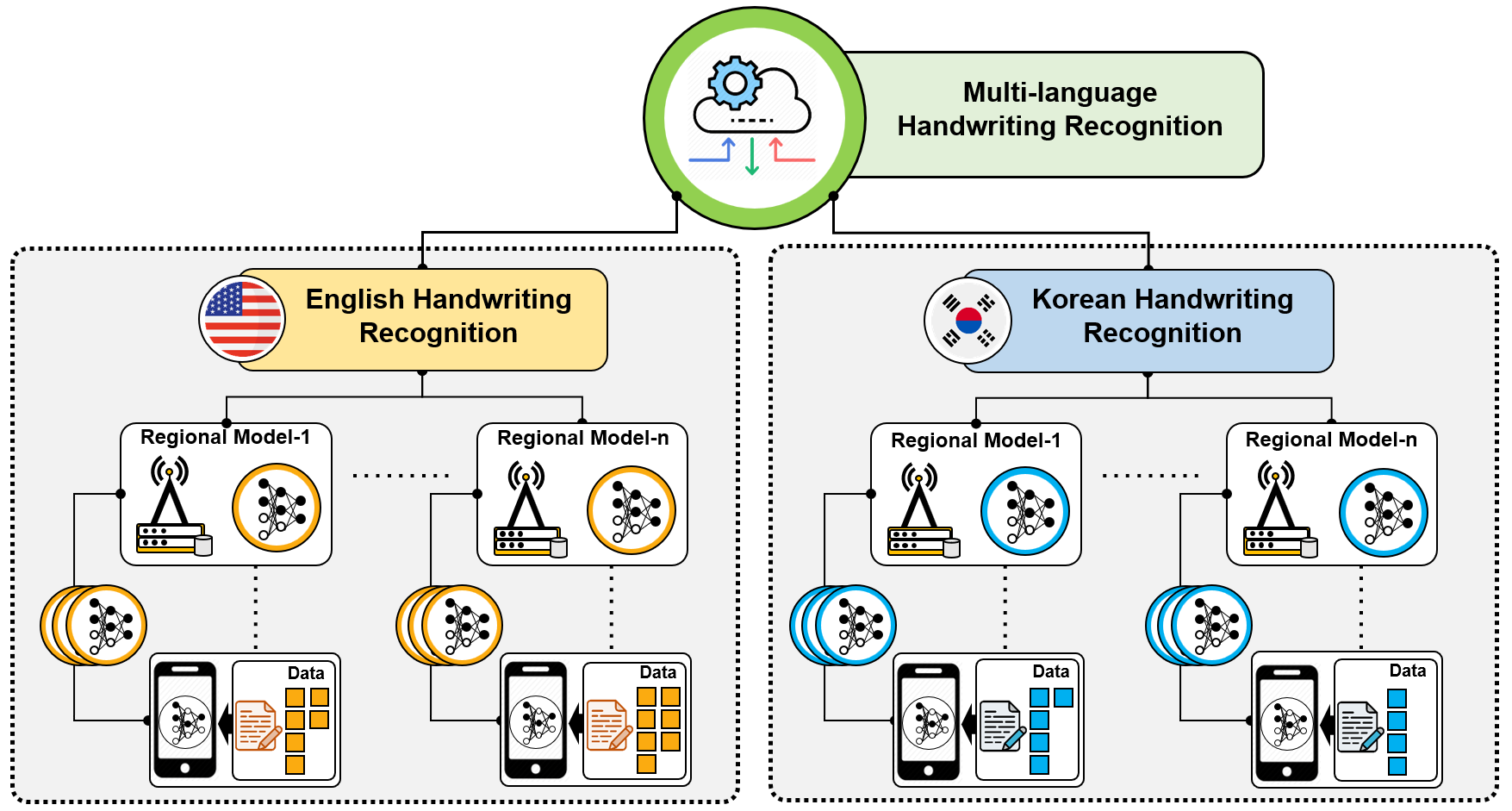}
		\caption{An example of Dem-AI systems: Multi-language handwriting recognition.}
		\label{F:Toy_Example}
	\end{figure*}

	\subsection{Generalization Mechanism}
	The generalization mechanism aims to collectively construct the hierarchical generalized knowledge from all existing specialized groups or learning agents, as illustrated in Fig. \ref{F:Generalization}. Accordingly, the generalized knowledge extends the generalization capability of the Dem-AI system to learn new tasks or deal with environment changes more efficiently.
	For this purpose, we propose four strategies that are suitable for different levels: \textit{direct knowledge sharing, indirect knowledge sharing, election,} and \textit{union}, which can be fixed or can be realized from a categorical distribution of these strategies. 
	At the lowest-level specialized groups, the direct knowledge sharing of learning agents is possible due to the similarity in the learning task to be performed. At the higher-level specialized groups, the indirect knowledge sharing among subgroups (i.e., transferred knowledge, meta-knowledge) becomes more probable due to the huge differences among specialized groups and the characteristics of learning tasks. Throughout the learning process, the groups become more and more specialized to efficiently solve different complex learning tasks. Consequently, the generalized knowledge of the specialized groups becomes very different at a higher level. Thus, an election mechanism based on voting can help in reaching consensus among specialized groups. To this end, a union mechanism is designed as an ensemble of the collection of highly-specialized groups. This is a possible way to maintain the diversity of potential groups for the entire learning system. Basically, the diversity of potential groups plays a vital role in the learning system. It allows the preservation of ineffective specialized groups who have fewer members or those who show low performance in the training setting, but are potentially able to deal with the changes in the training environment or new tasks. Therefore, the union and election mechanisms of Dem-AI are related to the diversity maintenance of the biological species through natural selection and non-competitive processes (i.e., symbiosis) in the evolution process or the robustness of decentralized systems \cite{johnson1999diversity}. In addition to the measurement of efficiency, the robustness or diversity of the Dem-AI system can be measured and controlled throughout the training time following a validation procedure.
	
	\textbf{Hierarchical learning model parameters averaging for knowledge sharing:}
	In the case of a shared generalized model among all group members, the direct knowledge sharing can be designed with the hierarchical averaging of the generalized model parameters (\textbf{GMP}) at each level $k$ as follows:
	
	\begin{align}
		\label{EQ:HierarchicalAvg}
		\textbf{GMP}(N_g^{(k)}) 
		&= (1-\gamma^t) \textbf{GMP}(N_g^{(k)}) \notag\\
		&+ \gamma^t \sum_{i \in \mathcal{S}} \frac{N_{g,i}^{(k-1)}}{N_g^{(k)}}\textbf{GMP}(N_{g,i}^{(k-1)}),
	\end{align}
	where $\mathcal{S}$ is the set of subgroups of a specialized group, $N_{g,i}^{(k-1)}$ is the number of agents in subgroup $i$, and  $N_{g}^{(k)}$ is the total number of agents in the current specialized group at level $k$. The model averaging implementation is a typical aggregation mechanism adopted in several FL algorithms \cite{FedProx2020, mcmahanCommunicationEfficientLearningDeep2017}. Parameter $\gamma^t$ controls the update frequency of the generalized knowledge whose value decreases over time as the members become well-specialized in their learning knowledge. Accordingly, the model parameters of the subgroups which have more numbers of agents become more important in the generalized model.

	\textbf{Knowledge distillation and knowledge transfer:}
	For multiple complex tasks, the Dem-AI framework allows knowledge transfer across tasks in different domains by leveraging collaboration amongst learning agents in the hierarchical structure. In this regard, multi-task learning enables generalization by solving multiple relevant tasks simultaneously \cite{ruder2017overview, zhang2017survey, standley2019tasks}. The work in \cite{standley2019tasks} studied the relationship between jointly trained tasks and proposed a framework for task grouping in MTL setting. Accordingly, the authors have analyzed learning task compatibility in computer vision systems by evaluating task cooperation and competition. For example, a shared encoder and representation is learned through training highly-correlated tasks together such as in Semantic Segmentation, Depth Estimation, and Surface Normal Prediction. However, this framework is limited to analyzing multiple learning tasks for a single agent. Whereas, in Dem-AI systems, a group of agents can train similar tasks in the low-level groups. Meanwhile, highly related tasks can be jointly trained together in the high-level groups.\par
	Furthermore, the latent representations across different devices or groups are supported by adopting existing techniques of knowledge distillation, transfer learning, meta-knowledge construction, and specialized knowledge transfer.
	Knowledge distillation \cite{park2019distilling} and knowledge transfer among multiple tasks \cite{mocha2017, corinzia2019variational} are important techniques to extend the capabilities of knowledge sharing. For example, in \cite{park2019distilling}, knowledge distillation mechanisms such as exchanging model parameters, model outputs, and surrogate data are incorporated in distributed machine learning frameworks. Meanwhile, knowledge transfer has been recently studied in the federated MTL setting using different types of MTL regularization such as cluster structure, probabilistic priors, and graphical models \cite{mocha2017}. Moreover, the work \cite{corinzia2019variational} forms a Bayesian network and uses variational inference methods with the lateral connections design between the server and client models  to transfer knowledge among tasks. Different from the recent works, the conventional \emph{organizational knowledge creation theory} \cite{org_knowledge1994} introduced a promising paradigm in which the new knowledge of an organization is articulated from the knowledge of individuals and self-organized in a hierarchical structure. Thus, the shared knowledge can be in an abstract form or an explicit combination of the individual's knowledge through the conceptualization and crystallization process. In doing so, together with the hierarchical learning model parameters averaging, we can develop suitable knowledge sharing approaches for the generalization mechanism in our Dem-AI systems. 
	
	
	\subsection{Example of Dem-AI Systems}
	More recently, the use of personalized applications, such as virtual assistants that could adhere to users' personality, has gained significant attraction. 
	The goal of such intelligent systems is to learn the unique features and personalized characteristics during daily activities and make appropriate decisions for each user, then enhance user interest. However, the main problem is the extraction of personalized features to perform knowledge transfer with limited local data. The Dem-AI system allows end-users and service providers to take part in a win-win solution, that is, the service providers exploit user's knowledge to scale up their services, and the end-users can collectively improve their personalized performance through knowledge sharing in a suitable group. For example, Google has provided a personalized virtual assistant (i.e., Google Now \cite{googleNow}) which can respond to user's questions with more relevant answers. Such reactive response systems can be extended to provide intelligent personalized recommendation services in a proactive manner. In this application, the hierarchical recommendation models can be constructed following Dem-AI mechanisms by leveraging the shared features from different domains, and users/groups at different levels.
	
	In addition, we present a novel multi-language handwriting recognition system based on our Dem-AI reference design, as shown in Fig. \ref{F:Toy_Example}. A typical handwriting language recognition application has an embedded virtual assistant to improve the capability for understanding human written texts in various languages. However, to realize such systems, we need separate recognition models for each language (e.g., English and Korean). Using our Dem-AI reference design, agents undergo self-organization to form appropriate hierarchical regional/social groups so as to share the similarity in the characteristics of their languages. By exploiting such structures, the learning system can collectively incorporate the personalized experiences of users that improve the generalization learning model. Subsequently, it empowers the recognition capability in each agent along with increasing the importance of the specialized process in the system. This kind of application can scale up to a large number of agents and support multiple languages. Thus, it has a potential to integrate different voice recognition models to develop a fully supporting virtual assistant for each client. Therefore, we unleash limitless possibilities for employing the Dem-AI philosophy in future distributed AI applications and, in the next section, we discuss new research challenges.
	
	\section{Extension of Existing Machine Learning Settings and Challenges}\label{S: Extension of Existing ML Settings and Challenges}
	The learning objective in the democratized learning setting for a large-scale distributed learning system cannot be readily solved by existing machine learning techniques. 
	Further, the limited design considerations and frameworks for both generalized and specialized capabilities of the distributed learning models necessitate a radical change in our approach to create efficient and better scalable learning systems. In the previous sections, as a first step to address these challenges, we established the Dem-AI philosophy and provided reference mechanisms from the interdisciplinary fields in nature. Accordingly, in the following subsections, we come up with new research challenges to develop future large-scale distributed learning systems that can leverage the Dem-AI philosophy, principle, and reference mechanisms.
	
	\subsection{Federated Learning towards Democratized Learning}
	Naturally, FL setting for a single learning task can be extended to multiple complex learning tasks with a very large number of biased learning agents in a democratized learning system. In addition, the learning agent's biases due to the characteristics of limited personalized data is a more general setting than the current non-i.i.d use cases of FL.
	Using the personalized learning problem \eqref{EQ:LPL_level_0}, the personalized model can incorporate the model of higher-level generalized groups by using proximal terms \cite{nguyen2020}. Also, the hierarchical averaging of learning model parameters in \eqref{EQ:HierarchicalAvg} could help agents to share their learning knowledge and construct more generalized knowledge of groups. Thereafter, the \emph{self-organizing hierarchical structuring} mechanism in Dem-AI can better adhere to the difference and similarity of the learning of the agents, which can be a promising direction to solve the problem of personalization and generalization more efficiently. 
	Moreover, Dem-AI also provides a better mechanism to handle newly arriving learning agents or deal with the changes in the agents (e.g., a change in their local datasets) due to the properties of self-organizing hierarchical structure and underlying dual processes. By moving new agents to suitable specialized groups, Dem-AI leverages new personalized knowledge for that group, where new members can also reuse the current specialized group knowledge. 
	
	
	In the current practical setting of the FL system, only a subset of available agents are chosen during the training process. This procedure leads to a very high-level group plasticity. Therefore, it is challenging to build a stable system having many levels of the hierarchical structure. In such a case, the number of levels in the hierarchical structure can be limited to a small number. The two essential corresponding research questions that can potentially revolutionize the FL towards Dem-AI systems are:
	\emph{1) How can we design a suitable self-organizing hierarchical structuring mechanism}, and \emph{2) How can we better leverage the generalized knowledge by using a new hierarchical averaging mechanism or other relevant sharing knowledge approaches.} 
	In addition, we can extend the Dem-AI philosophy to other distributed learning systems that are analogous to FL, such as the brainstorming generative adversarial network (GAN) system proposed in ~\cite{ferdowsi2020brainstorming} which applies FL-like principles to generative models rather than inference models. 
	
	\subsection{Cooperative Multi-Agent Reinforcement Learning}
	The setting of cooperative MARL discussed in \cite{NIPS2019_multiagent} requires a coordination of decentralized policies for solving complex tasks due to the partial observability $\boldsymbol{o}_n$ of each agent $n$ (e.g., different fields of view). In fact, shared common knowledge and hierarchical policy design arise naturally in the decentralized cooperative tasks, as discussed in \cite{NIPS2019_multiagent}. 
	A cooperative reward $r(\boldsymbol{s},\boldsymbol{a}_{\textrm{joint}})$ is built according to the function of the joint action (i.e.,
	$\boldsymbol{a}_{\textrm{joint}}\eqdef \{ a_1, \dots, a_n\}$). 
	Several other approaches such as centralized critic, decentralized execution in MADDPG \cite{ma_ddpg2017}, hierarchical critic in \cite{cao2019reinforcement}, and feudal framework \cite{ahilan2019feudal} introduce various designs of decentralized operation in cooperative reinforcement learning. However, these frameworks mainly analyze two levels of the hierarchical structure with a small number of agents that could be suitable only for a group learning.
	As a result, scaling up the design with a large number of learning agents who need to perform multiple coordination tasks would be very challenging using the existing designs. Existing federated reinforcement learning designs \cite{nadiger2019federated,wang2020federated} can be readily incorporated with our mechanisms, as discussed in prior sections. Nevertheless, in order to fully realize Dem-AI principles and mechanisms, we must overcome two key challenges: a) how to develop novel similarity metrics for group formation (e.g., observations/tasks/goals based metrics), and b) how to realize suitable multi-level cooperation for knowledge acquisition among groups of agents.
	
	
	Furthermore, current MARL systems are ineptly designed for handling environmental changes, such as the deployment of new agents or the deployment in different environments. We believe that the democratized learning philosophy and the presented reference mechanisms can provide more flexible approaches to control exploration and exploitation capabilities with a self-organizing hierarchical structure of the agents. This can help MARL systems as those in \cite{NIPS2019_multiagent} and \cite{ma_ddpg2017} evolve towards a better scalable and powerful design. Dem-AI also provides an opportunity to collectively train each agent for multiple basic DRL tasks, and then the collective knowledge can help to solve more difficult tasks with specialized groups of agents. Subsequently, such decentralized autonomous systems can be widely applied to handle multiple complex tasks in future applications. In our recent work \cite{kasgari2019experienced}, we showed a simple Dem-AI principle for DRL whereby we allowed a DRL agent to gain ``experience'' on extreme events (which can be seen as a specialized process) by training over a GAN-based system. Using this early work, one can build a more elaborate MA-DRL system under the Dem-AI umbrella.
	
	\subsection{Multi-Task Learning and Meta-Learning}
	The current setting of MTL and meta-learning is restricted to train similar tasks or strongly correlated learning tasks, and it focuses on maintaining a certain level of generalization and performance for each learning task. Recent federated MTL frameworks such as those in \cite{mocha2017} and \cite{corinzia2019variational} resolve the statistical challenges of different user-dependent data distributions in the classical FL with different but similar learning tasks for each client. A general formulation in \cite{mocha2017} introduced a trainable correlation matrix $\boldsymbol{\Omega}$ between tasks as follows:
	\begin{align}
		\underset{\boldsymbol{w}, \boldsymbol{\Omega}}{\min}~ \sum_n L_n(\boldsymbol{w}|\mathcal{D}^{(0)}_n) + R(\boldsymbol{w}, \boldsymbol{\Omega}),
	\end{align}
	where a variety of regularizer function $R$ can be implemented for the clustered multi-task learning problem \cite{zhou2011clustered, zhang2017survey}. Thereafter, we can incorporate the proximal terms in \eqref{EQ:LPL_level_0} and hierarchical averaging in \eqref{EQ:HierarchicalAvg} into this federated MTL framework and build a hierarchical generalization by grouping similar or strongly correlated tasks.
	
	Similarly, to turn the existing federated MLF frameworks in \cite{chen2018federated, fallah2020personalized} into practical large-scale distributed learning systems, Dem-AI mechanisms are also necessary to efficiently deal with new tasks which can join a suitable correlated group of tasks, eliminating the need for them to be very similar to all of the training tasks. Therefore, grouping a large number of tasks in a self-organizing hierarchical structure with different levels of knowledge transfer learning is one of the promising designs to extend MLF and MTL frameworks towards a large-scale design.

	\subsection{Transfer Learning}
	Transfer learning is an important technique that can help in the sharing of knowledge among specialized group members and among multiple tasks in democratized learning. The generalized knowledge can be transferable, directly or indirectly, due to the similar characteristic such as in-group sharing based on recent works: federated distillation \cite{park2019distilling} or novel GAN designs for distributed datasets \cite{Dec_GAN2019,ferdowsi2020brainstorming}. However, the high level of specialized knowledge is often difficult to transfer due to high dissimilarity and incompatibility of the knowledge. Therefore, we need a novel approach to extract specialized knowledge from different learning tasks. Moreover, a consensus of incompatible knowledge can be reached among the members or specialized groups by the novel election and union mechanism. Therefore, the hierarchical structure in transfer learning is a promising research direction that will enable the democratized learning system to solve multiple complex learning tasks more efficiently.
	
	\section{Conclusion and Future Works}\label{S:Conclusion}
	Existing machine learning designs face critical challenges in scaling up the current centralized AI systems into the distributed AI systems that can perform multiple complex learning tasks. In this paper, we have established the principles of a novel democratized learning setting, dubbed as Dem-AI, while reviewing and incorporating the natural design considerations for the distributed machine learning systems. As an initial step towards this, we have first established a natural design approach using the Dem-AI philosophy and its reference mechanisms from various interdisciplinary fields designed for the large-scale distributed learning systems. In particular, we have presented the evolution of the specialized and generalized processes and the formation of self-organizing hierarchical structure in the Dem-AI principle. Next, with Dem-AI, we have introduced possible extensions on machine learning settings and new challenges for the existing learning approaches to provide better scalable and flexible learning systems. The effects of transitions in Dem-AI principle should be further validated for different specific learning settings and applications. We leave the validation analysis of the proposed Dem-AI principle for our future works.

	
	%

	%

	
	\section*{Acknowledgment}
	
	This work was supported by the National Research Foundation of Korea (NRF) grant funded by the Korea government (MSIT) (No.2020R1A4A1018607).

	\ifCLASSOPTIONcaptionsoff
	\newpage
	\fi

	
	
	%
	%
	%
	
	\bibliographystyle{IEEEtran}
	\bibliography{Dem-AI}

\begin{thebibliography}{10}
\providecommand{\url}[1]{#1}
\csname url@samestyle\endcsname
\providecommand{\newblock}{\relax}
\providecommand{\bibinfo}[2]{#2}
\providecommand{\BIBentrySTDinterwordspacing}{\spaceskip=0pt\relax}
\providecommand{\BIBentryALTinterwordstretchfactor}{4}
\providecommand{\BIBentryALTinterwordspacing}{\spaceskip=\fontdimen2\font plus
\BIBentryALTinterwordstretchfactor\fontdimen3\font minus
  \fontdimen4\font\relax}
\providecommand{\BIBforeignlanguage}[2]{{%
\expandafter\ifx\csname l@#1\endcsname\relax
\typeout{** WARNING: IEEEtran.bst: No hyphenation pattern has been}%
\typeout{** loaded for the language `#1'. Using the pattern for}%
\typeout{** the default language instead.}%
\else
\language=\csname l@#1\endcsname
\fi
#2}}
\providecommand{\BIBdecl}{\relax}
\BIBdecl

\bibitem{Chen_tutorial}
M.~{Chen}, U.~{Challita}, W.~{Saad}, C.~{Yin}, and M.~{Debbah}, ``Artificial
  neural networks-based machine learning for wireless networks: A tutorial,''
  \emph{IEEE Communications Surveys Tutorials}, vol.~21, no.~4, pp. 3039--3071,
  Fourthquarter 2019.

\bibitem{FL_advances}
P.~Kairouz, H.~B. McMahan, B.~Avent, A.~Bellet, M.~Bennis, A.~N. BhagCanhoji,
  K.~Bonawitz, Z.~Charles, G.~Cormode, R.~Cummings \emph{et~al.}, ``Advances
  and open problems in federated learning,'' \emph{arXiv:1912.04977}, 2019.

\bibitem{tran2019federated}
N.~H. {Tran}, W.~{Bao}, A.~{Zomaya}, M.~N.~H. {Nguyen}, and C.~S. {Hong},
  ``Federated learning over wireless networks: Optimization model design and
  analysis,'' in \emph{IEEE Conference on Computer Communications (INFOCOM)},
  Paris, France, April 29-- May 2 , 2019, pp. 1387--1395.

\bibitem{pandey2020crowdsourcing}
S.~R. Pandey, N.~H. Tran, M.~Bennis, Y.~K. Tun, A.~Manzoor, and C.~S. Hong, ``A
  crowdsourcing framework for on-device federated learning,'' \emph{IEEE
  Transactions on Wireless Communications}, 2020.

\bibitem{FL_TON2019}
C.~Dinh, N.~H. Tran, M.~N.~H. Nguyen, C.~S. Hong, W.~Bao, A.~Y. Zomaya, and
  V.~Gramoli, ``{Federated Learning over Wireless Networks: Convergence
  Analysis and Resource Allocation},'' \emph{arXiv:1910.13067}, 2019.

\bibitem{khan2019federated}
L.~U. Khan, N.~H. Tran, S.~R. Pandey, W.~Saad, Z.~Han, M.~N.~H. Nguyen, and
  C.~S. Hong, ``Federated learning for edge networks: Resource optimization and
  incentive mechanism,'' \emph{arXiv:1911.05642}, 2019.

\bibitem{chen2019joint}
M.~Chen, Z.~Yang, W.~Saad, C.~Yin, H.~V. Poor, and S.~Cui, ``A joint learning
  and communications framework for federated learning over wireless networks,''
  \emph{arXiv:1909.07972}, 2019.

\bibitem{chen2020convergence}
M.~Chen, H.~V. Poor, W.~Saad, and S.~Cui, ``Convergence time optimization for
  federated learning over wireless networks,'' \emph{arXiv preprint
  arXiv:2001.07845}, 2020.

\bibitem{maml2017}
C.~Finn, P.~Abbeel, and S.~Levine, ``Model-agnostic meta-learning for fast
  adaptation of deep networks,'' in \emph{Proceedings of the 34th International
  Conference on Machine Learning}, Sydney, NSW, Australia, Aug. 2017, pp.
  1126–--1135.

\bibitem{mocha2017}
V.~Smith, C.-K. Chiang, M.~Sanjabi, and A.~S. Talwalkar, ``Federated multi-task
  learning,'' in \emph{Advances in Neural Information Processing Systems 30},
  Dec. 2017, pp. 4424--4434.

\bibitem{corinzia2019variational}
L.~Corinzia and J.~M. Buhmann, ``Variational federated multi-task learning,''
  \emph{arXiv:1906.06268}, 2019.

\bibitem{mcmahanCommunicationEfficientLearningDeep2017}
B.~McMahan, E.~Moore, D.~Ramage, S.~Hampson, and B.~A. y~Arcas,
  ``Communication-efficient learning of deep networks from decentralized
  data,'' in \emph{Artificial Intelligence and Statistics}, 2017, pp.
  1273--1282.

\bibitem{FedProx2020}
T.~Li, A.~K. Sahu, M.~Zaheer, M.~Sanjabi, A.~Talwalkar, and V.~Smith,
  ``Federated optimization in heterogeneous networks,'' in \emph{Proceedings of
  Machine Learning and Systems 2020}, 2020, pp. 429--450.

\bibitem{fallah2020personalized}
A.~Fallah, A.~Mokhtari, and A.~Ozdaglar, ``Personalized federated learning: A
  meta-learning approach,'' \emph{arXiv:2002.07948}, 2020.

\bibitem{mansour2020three}
Y.~Mansour, M.~Mohri, J.~Ro, and A.~T. Suresh, ``Three approaches for
  personalization with applications to federated learning,'' \emph{arXiv
  preprint arXiv:2002.10619}, 2020.

\bibitem{LifeLong2019}
G.~I. Parisi, R.~Kemker, J.~L. Part, C.~Kanan, and S.~Wermter, ``Continual
  lifelong learning with neural networks: A review,'' \emph{Neural Networks},
  vol. 113, pp. 54--71, 2019.

\bibitem{oudeyer2018computational}
P.-Y. Oudeyer, ``Computational theories of curiosity-driven learning,''
  \emph{arXiv:1802.10546}, 2018.

\bibitem{brambilla2013swarm}
M.~Brambilla, E.~Ferrante, M.~Birattari, and M.~Dorigo, ``Swarm robotics: a
  review from the swarm engineering perspective,'' \emph{Swarm Intelligence},
  vol.~7, no.~1, pp. 1--41, 2013.

\bibitem{NIPS2019_multiagent}
C.~Schroeder~de Witt, J.~Foerster, G.~Farquhar, P.~Torr, W.~Boehmer, and
  S.~Whiteson, ``Multi-agent common knowledge reinforcement learning,'' in
  \emph{Advances in Neural Information Processing Systems 32}.\hskip 1em plus
  0.5em minus 0.4em\relax Curran Associates, Inc., 2019, pp. 9927--9939.

\bibitem{ruder2017overview}
S.~Ruder, ``An overview of multi-task learning in deep neural networks.''

\bibitem{zhang2017survey}
Y.~Zhang and Q.~Yang, ``A survey on multi-task learning,''
  \emph{arXiv:1707.08114}, 2017.

\bibitem{park2019distilling}
J.~Park, S.~Wang, A.~Elgabli, S.~Oh, E.~Jeong, H.~Cha, H.~Kim, S.-L. Kim, and
  M.~Bennis, ``Distilling on-device intelligence at the network edge,''
  \emph{arXiv:1908.05895}, 2019.

\bibitem{H_Clustering}
G.~Karypis, E.~H. Han, and V.~Kumar, ``{Chameleon: Hierarchical clustering
  using dynamic modeling},'' \emph{Computer}, vol.~32, no.~8, pp. 68--75, Aug.
  1999.

\bibitem{zurn2020networkhierarchy}
P.~Zurn and D.~S. Bassett, ``Network architectures supporting learnability,''
  \emph{Philosophical Transactions of the Royal Society B}, vol. 375, no. 1796,
  p. 20190323, 2020.

\bibitem{mengistu2016evolutionaryhierarchy}
H.~Mengistu, J.~Huizinga, J.-B. Mouret, and J.~Clune, ``The evolutionary
  origins of hierarchy,'' \emph{PLoS computational biology}, vol.~12, no.~6, p.
  e1004829, 2016.

\bibitem{pendulum}
S.~J. Ling, J.~Sanny, W.~Moebs \emph{et~al.}, \emph{University Physics Volume
  1}.\hskip 1em plus 0.5em minus 0.4em\relax OpenStax, 2016.

\bibitem{sattler2020clustered}
F.~Sattler, K.-R. M{\"u}ller, and W.~Samek, ``Clustered federated learning:
  Model-agnostic distributed multitask optimization under privacy
  constraints,'' \emph{IEEE Transactions on Neural Networks and Learning
  Systems}, pp. 1--13, 2020.

\bibitem{han_niyato_saad_basar_2019}
Z.~Han, D.~Niyato, W.~Saad, and T.~Ba\c{s}ar, \emph{Game Theory for Next
  Generation Wireless and Communication Networks: Modeling, Analysis, and
  Design}.\hskip 1em plus 0.5em minus 0.4em\relax Cambridge University Press,
  2019.

\bibitem{hierarchicalgame_2014}
L.~{Rose}, E.~V. {Belmega}, W.~{Saad}, and M.~{Debbah}, ``Pricing in
  heterogeneous wireless networks: Hierarchical games and dynamics,''
  \emph{IEEE Transactions on Wireless Communications}, vol.~13, no.~9, pp.
  4985--5001, Sep. 2014.

\bibitem{coalitionalgame}
W.~Saad, Z.~Han, M.~Debbah, A.~Hj{\o}rungnes, and T.~Ba\c{s}ar, ``Coalitional
  game theory for communication networks,'' \emph{IEEE Signal Processing
  Magazine}, vol.~26, no.~5, pp. 77--97, Sep. 2009.

\bibitem{kornblith2019similarity}
S.~Kornblith, M.~Norouzi, H.~Lee, and G.~Hinton, ``Similarity of neural network
  representations revisited,'' \emph{arXiv:1905.00414}, 2019.

\bibitem{iannello2013immune}
A.~Iannello and D.~H. Raulet, ``Immune surveillance of unhealthy cells by
  natural killer cells,'' in \emph{Cold Spring Harbor symposia on quantitative
  biology}, vol.~78, 2013, pp. 249--257.

\bibitem{nguyen2020}
M.~N.~H. Nguyen, S.~R. Pandey, T.~Nguyen~D., E.~N. Huh, C.~S. Hong, N.~H. Tran,
  and W.~Saad, ``Self-organizing democratized learning: Towards large-scale
  distributed learning systems,'' \emph{arXiv:2007.03278}, 2020.

\bibitem{sattler2019robust}
F.~Sattler, S.~Wiedemann, K.-R. M{\"u}ller, and W.~Samek, ``Robust and
  communication-efficient federated learning from non-iid data,'' \emph{IEEE
  transactions on neural networks and learning systems}, 2019.

\bibitem{wang2019edgeIntelligence}
X.~Wang, Y.~Han, C.~Wang, Q.~Zhao, X.~Chen, and M.~Chen, ``In-edge ai:
  Intelligentizing mobile edge computing, caching and communication by
  federated learning,'' \emph{IEEE Network}, vol.~33, no.~5, pp. 156--165,
  2019.

\bibitem{ma_ddpg2017}
R.~Lowe, Y.~Wu, A.~Tamar, J.~Harb, P.~Abbeel, and I.~Mordatch, ``Multi-agent
  actor-critic for mixed cooperative-competitive environments,'' in
  \emph{Advances in Neural Information Processing Systems 30}, Dec. 2017, pp.
  6379--6390.

\bibitem{johnson1999diversity}
N.~L. Johnson, ``Diversity in decentralized systems: Enabling self-organizing
  solutions,'' in \emph{Decentralization II Conference, UCLA}, 1999.

\bibitem{standley2019tasks}
T.~Standley, A.~R. Zamir, D.~Chen, L.~Guibas, J.~Malik, and S.~Savarese,
  ``Which tasks should be learned together in multi-task learning?''
  \emph{arXiv:1905.07553}, 2019.

\bibitem{org_knowledge1994}
I.~Nonaka, ``A dynamic theory of organizational knowledge creation,''
  \emph{Organization science}, vol.~5, no.~1, pp. 14--37, 1994.

\bibitem{googleNow}
``Google now,'' https://www.wordstream.com/google-now.

\bibitem{ferdowsi2020brainstorming}
A.~Ferdowsi and W.~Saad, ``Brainstorming generative adversarial networks
  ({BGANs}): Towards multi-agent generative models with distributed private
  datasets,'' \emph{arXiv:2002.00306}, 2020.

\bibitem{cao2019reinforcement}
Z.~Cao and C.-T. Lin, ``Reinforcement learning from hierarchical critics,''
  \emph{arXiv:1902.03079}, 2019.

\bibitem{ahilan2019feudal}
S.~Ahilan and P.~Dayan, ``Feudal multi-agent hierarchies for cooperative
  reinforcement learning,'' \emph{arXiv:1901.08492}, 2019.

\bibitem{nadiger2019federated}
C.~Nadiger, A.~Kumar, and S.~Abdelhak, ``Federated reinforcement learning for
  fast personalization,'' in \emph{2019 IEEE Second International Conference on
  Artificial Intelligence and Knowledge Engineering (AIKE)}.\hskip 1em plus
  0.5em minus 0.4em\relax IEEE, 2019, pp. 123--127.

\bibitem{wang2020federated}
X.~Wang, C.~Wang, X.~Li, V.~C. Leung, and T.~Taleb, ``Federated deep
  reinforcement learning for internet of things with decentralized cooperative
  edge caching,'' \emph{IEEE Internet of Things Journal}, 2020.

\bibitem{kasgari2019experienced}
A.~T.~Z. Kasgari, W.~Saad, M.~Mozaffari, and H.~V. Poor, ``Experienced deep
  reinforcement learning with generative adversarial networks ({GANs}) for
  model-free ultra reliable low latency communication,''
  \emph{arXiv:1911.03264}, 2019.

\bibitem{zhou2011clustered}
J.~Zhou, J.~Chen, and J.~Ye, ``Clustered multi-task learning via alternating
  structure optimization,'' in \emph{Advances in neural information processing
  systems}, 2011, pp. 702--710.

\bibitem{chen2018federated}
F.~Chen, M.~Luo, Z.~Dong, Z.~Li, and X.~He, ``Federated meta-learning with fast
  convergence and efficient communication,'' \emph{arXiv:1802.07876}, 2018.

\bibitem{Dec_GAN2019}
C.~{Hardy}, E.~{Le Merrer}, and B.~{Sericola}, ``Md-gan: Multi-discriminator
  generative adversarial networks for distributed datasets,'' in \emph{2019
  IEEE International Parallel and Distributed Processing Symposium (IPDPS)},
  Brazil, May 2019, pp. 866--877.

\end{thebibliography}
	
	%
	
	%
	%
	%
	
	
	

\end{document}